\documentclass[conference]{IEEEtran}

\usepackage[T1]{fontenc}  

\usepackage[utf8]{inputenc}
\usepackage{graphicx}
\usepackage{subcaption}

\RequirePackage[pdftex,hidelinks]{hyperref}
\usepackage{outlines}
\usepackage[shortlabels]{enumitem}
\usepackage[pdftex,dvipsnames,table,xcdraw]{xcolor}
\usepackage{graphicx}

\usepackage{booktabs}
\usepackage{multirow}
\usepackage{graphicx}
\usepackage{multirow}
\usepackage{multicol}
\usepackage{tabularx}

\usepackage{amsmath}

\DeclareRobustCommand{\IEEEauthorrefmark}[1]{\smash{\textsuperscript{\footnotesize #1}}}

\begin{document}

\title{
Evaluating GPT-3.5 and GPT-4 Models on Brazilian University Admission Exams}

\author{\IEEEauthorblockN{Desnes Nunes\IEEEauthorrefmark{*1},
Ricardo Primi\IEEEauthorrefmark{*2}, Ramon Pires\IEEEauthorrefmark{*3},
Roberto Lotufo\IEEEauthorrefmark{4,5}, and
Rodrigo Nogueira\IEEEauthorrefmark{3,4,5}}
\\
\IEEEauthorblockA{
\IEEEauthorrefmark{1}University of São Paulo (USP),
\IEEEauthorrefmark{2}University of San Francisco (USF)\\
\IEEEauthorrefmark{3}Maritaca AI,
\IEEEauthorrefmark{4}NeuralMind,
\IEEEauthorrefmark{5}State University of Campinas (UNICAMP)\\
Brazil\\
\IEEEauthorrefmark{*}Equal contribution
}}

\maketitle

\begin{abstract}
The present study aims to explore the capabilities of Language Models (LMs) in tackling high-stakes multiple-choice tests, represented here by the \textit{Exame Nacional do Ensino Médio} (ENEM), a multidisciplinary entrance examination widely adopted by Brazilian universities. This exam poses challenging tasks for LMs, since its questions may span into multiple fields of knowledge, requiring understanding of information from diverse domains. For instance, a question may require comprehension of both statistics and biology to be solved. This work analyzed responses generated by GPT-3.5 and GPT-4 models for questions presented in the 2009-2017 exams, as well as for questions of the 2022 exam, which were made public after the training of the models was completed. Furthermore, different prompt strategies were tested, including the use of Chain-of-Thought (CoT) prompts to generate explanations for answers. On the 2022 edition, the best-performing model, GPT-4 with CoT, achieved an accuracy of 87\%, largely surpassing GPT-3.5 by 11 points.
The code and data used on experiments are available at \url{https://github.com/piresramon/gpt-4-enem}
\end{abstract}

\section{Introduction}
Over the past decade, advancements in natural language processing (NLP) and machine learning (ML) have led to the development of increasingly sophisticated Language Models (LMs). These models have demonstrated remarkable performance on a wide array of tasks, such as translation, summarization, question-answering, among many others~\cite{devlin2018bert,brown2020language,rae2022scaling,chowdhery2022palm,hoffmann2022training,touvron2023llama}. As a consequence, the education sector has begun to explore the potential benefits of these technologies, with several studies investigating their potential in classrooms \cite{kuvcak2018machine} and the concept of ``precision education'' \cite{luan2021review}.

However, most existing research focuses on evaluating LMs in the context of English language tasks, with some notable exceptions~\cite{ahuja2023mega,BERTIN-GPT,zeng2022glm130b,xue2021mt5,xue2022byt5,scao2022bloom,muennighoff2022crosslingual}. In particular, there is a lack of studies examining performance of LMs on Portuguese tasks.

In this work, we evaluate state-of-the-art LMs on \textit{Exame Nacional do Ensino Médio} (\href{https://www.gov.br/inep/pt-br/areas-de-atuacao/avaliacao-e-exames-educacionais/enem}{\underline{ENEM}}), a multidisciplinary admission test widely used in Brazilian universities.
The ENEM exam, presented in Brazilian Portuguese, poses a unique challenge for LMs, as it requires a deep understanding of various fields of knowledge and the ability to integrate information from diverse domains. We analyze responses generated by the latest GPT-3.5 and GPT-4 models to questions of the ENEM exams using different prompt strategies, including zero-shot and few-shot prompts with Chain-of-Thought (CoT) explanations. Moreover, to tackle the potential issue of the model memorizing the answers to the questions during training, an evaluation was performed on the 2022 exam, which was made public subsequent to the completion of their training.

By assessing the performance of the state-of-the-art GPT models in this non-English, high-stakes exam context, this study contributes to the growing body of research exploring the applicability of LMs in educational settings. Furthermore, the findings of this study have the potential to inform the development of future LMs that are better equipped to handle diverse, multidisciplinary tasks in various languages, ultimately promoting a more inclusive and accessible landscape for AI-driven educational tools.

\section{Related Work}

The goal of this research is to promote advances and evaluate the performance of automatic question resolution of college entrance exams utilizing recent state-of-the-art LMs. The multiple-choice questions presented in the ENEM examination pose a complex challenge, requiring advanced NLP techniques to be solved. Similar to the Scholastic Assessment Test (\href{https://collegereadiness.collegeboard.org/sat}{\underline{SAT}}), ENEM is an exam that assesses the knowledge of students who intend to enter universities in Brazil.\footnote{The exam is majorly divided into four areas: languages, codes and their technologies; human sciences and their technologies; natural sciences and their technologies; mathematics and its technologies. Moreover, the exam is composed of 180 multiple-choice questions and an essay.} A prior study of AI models applied to ENEM~\cite{silveira2018enem} tackled this challenge by employing static Word Embeddings~\cite{MikolovEfficientEstimationWord2013a} and WordNet~\cite{miller1995wordnet}, but attained limited results, ranging between 26-29\% on accuracy. This current paper revisits the task, leveraging the zero-shot and few-shot capabilities of GPT-3.5 and 4 models.

A recent study applied GPT-3.5 to the United States Bar Examination, achieving a passing performance in two out of seven categories~\cite{bommarito2022gpt}. In a separate study, it was discovered that GPT-3.5 falls significantly short of human performance in analytical quantitative reasoning questions on the Certified Public Accountants (CPA) Examination~\cite{Bommarito2023Gpt}. Nevertheless, it demonstrates a comparable level of performance to humans in questions that demand the skills of remembering, understanding, and applying knowledge.

In the medical domain, researchers finetuned PALM~\cite{chowdhery2022palm}, a large LM pretrained on diverse texts, on a curated set of medical-related question-answering examples~\cite{singhal2022large}. The resulting model, Med-PALM, was evaluated on questions from the United States Medical Licensing Examination (USMLE). Their analysis revealed that the model provided answers in agreement with the scientific consensus, as determined by clinical experts, for 92.6\% of the questions.

More recently, GPT-4~\cite{openai2023gpt4} demonstrated performance comparable to humans across multiple professional and academic benchmarks, such as achieving a score within the top 10\% of participants on a simulated bar exam. It also largely surpasses Med-PALM on a version of the USMLE benchmark~\cite{nori2023capabilities}. We are the first to evaluate this model on a Portuguese benchmark.

\section{Methodology}

\subsection{Datasets}

We use two evaluation datasets: The ENEM Challenge and ENEM 2022.
The ENEM Challenge dataset\footnote{\url{https://www.ime.usp.br/~ddm/project/enem}}~\cite{silveira2018enem} was created by parsing questions and alternatives from several editions of the annual ENEM examination. The dataset's authors have also annotated each question with the following tags indicating the domain:

\begin{itemize}
  \item Text Comprehension (TC)
  \item Encyclopedic Knowledge (EK)
  \item Image Comprehension (IC)
  \item Domain Speciﬁc Knowledge (DS)
  \item Mathematical Reasoning (MR)
  \item Chemical Elements (CE)
\end{itemize}

This informative knowledge tag labeling is crucial to determine whether a question contains elements that cannot be treated as text, such as images or chemical symbols. The complete dataset comprises 1754 ENEM questions spanning from 2009-2017. Following~\cite{silveira2018enem}, we eliminated questions requiring image comprehension (IC), mathematical reasoning (MR), and having chemical elements (CE), resulting in 916 questions. Even though the evaluated models possess the means to process equations or chemical symbols, a comprehensive analysis of these aspects has been limited to the 2022 exam that was parsed and annotated by us. The questions in the ENEM Challenge dataset used in this study are composed of three components: a \textbf{header}, which presents the main text; a \textbf{statement}, which poses the question to the students; and \textbf{alternatives}, which provide five options for the student to choose the correct answer to the statement.

In addition, we evaluated the models on ENEM 2022 questions, which is the most recent edition of the exam. The dataset was structured and annotated similarly to the ENEM Challenge dataset. Table~\ref{tab:stats_enem_2022} presents the statistics of the ENEM 2022 dataset. Using the same filtering criteria applied to the ENEM Challenge dataset, ENEM 2022 yields 88 questions. By only removing questions that require image comprehension (IC), we have 118 questions.
As for the original ENEM Challenge dataset, here we also exclude the five questions related to the foreign language (English or Spanish) chosen at the time of registration.

\begin{table}[]
\centering
\caption{Number of questions of ENEM 2022 dataset.}
\begin{tabular}{@{}lccc@{}}
\toprule
Area                                        & total  & \tiny{$\neg$ IC $\land$ $\neg$ MR $\land$ $\neg$ CE }  & \tiny{$\neg$ IC}  \\ \midrule
Languages, Codes and their Technologies     & 45     & 33                              & 33  \\
Human Sciences and their Technologies       & 45     & 37                              & 37  \\
Natural Sciences and their Technologies     & 45     & 18                              & 26  \\ 
Mathematics and its Technologies            & 45     & 0                               & 22  \\ \midrule
Total                                       & 180    & 88                              & 118  \\ \bottomrule
\end{tabular}%
\label{tab:stats_enem_2022}
\end{table}

\subsection{The GPT Models}

In this study, we use OpenAI's GPT 3.5 and 4 models, which are available via a paid API.\footnote{\url{https://platform.openai.com}} The models we evaluated include \texttt{code-davinci-002}, \texttt{text-davinci-002}, \texttt{text-davinci-003}, \texttt{gpt-3.5-turbo-0314}, and \texttt{gpt-4-0314}. Except for \texttt{gpt-4-0314}, these models fall under the GPT-3.5 family, as designated by OpenAI. For all models, we used the following parameters: temperature of 0, which defaults to greedy decoding, maximum output of 4000 tokens, frequency penalty of 0, and presence penalty of 0.

We do not know the inner workings of the models, save for a blog post detailing certain distinctions in their training methodology.\footnote{\url{https://platform.openai.com/docs/model-index-for-researchers}} We know however that they were exclusively trained on data publicly available until September 2021.\footnote{\url{https://platform.openai.com/docs/models}} Hence, the results in the ENEM 2022 test are trustworthy, given that the models could not have access to the examination questions and answers during their training phase.\footnote{The ENEM 2022 exam was administered on November 13-20, 2022.}

\subsection{Prompt Design}

We used three types of prompts to analyze multiple-choice questions: zero-shot, few-shot, and few-shot with CoT. Each prompt is described hereafter.

\subsubsection{\textbf{Zero-shot}}

For zero-shot prompts, the question is initially parsed into a pre-defined prompt format and then inputted into the model, which then generates an answer. Figure~\ref{fig:example_enem_2009} illustrates a zero-shot prompt example (translated into English).

\begin{figure}[!htb]
\sffamily
\scriptsize
\centering
\begin{tabular}{|
>{\columncolor[HTML]{ECF4FF}}p{0.9\columnwidth} |}
\hline
{\color[HTML]{343434} \textbf{Question 1:}}  \\
{\color[HTML]{343434} \textbf{Header:} The biogeochemical cycle of carbon comprises several compartments, including the Earth, atmosphere and oceans, and several processes that allow the transfer of compounds between these reservoirs. Carbon stocks stored in the form of non-renewable resources, for example oil, are limited, and it is of great importance to realize the importance of replacing fossil fuels with fuels from renewable sources.} \\
{\color[HTML]{343434} \textbf{Statement:} The use of fossil fuels interferes with the carbon cycle, as it causes} \\
{\color[HTML]{343434} \textbf{Alternatives:}}     \\ 
{\color[HTML]{343434} A. Increase in the percentage of carbon contained in the Earth.} \\ 
{\color[HTML]{343434} B. reduction in the rate of photosynthesis of higher plants.} \\ 
{\color[HTML]{343434} C. increased production of plant-based carbohydrates.} \\ 
{\color[HTML]{343434} D. increase in the amount of carbon present in the atmosphere.} \\ 
{\color[HTML]{343434} E. Reducing the global amount of carbon stored in the oceans.} \\ 
{\color[HTML]{343434} \textbf{Answer:}} {\color[HTML]{00009B} \textit{D. increase in the amount of carbon present in the atmosphere.}}     \\ \hline
\end{tabular}
\caption{Example of zero-shot prompt for question 6 from the ENEM 2009 exam. The text in blue is the output generated by the model.}
\label{fig:example_enem_2009}
\end{figure}

By not having access to few-shot examples, the model tends to produce responses with format variations (e.g., ``D'', ``D. increase in the amount of carbon present in the atmosphere'', or ``Alternative D.''). In preliminary experiments, we observed that the use of instructions was not effective in establishing the desired output format. To overcome this challenge, we have developed a set of regex rules to identify the chosen alternative (A, B, C, D, or E) and filter out the irrelevant text additions.

\subsubsection{\textbf{Few-shot}}
\label{sec:fewshot_examples}

The few-shot prompt is composed of some examples that induce the model to generate responses in the expected format.

We selected three examples of different knowledge areas from ENEM 2022 as our few-shot examples. Specifically, we chose one question from ``Languages, Codes and their Technologies", one from ``Human Sciences and their Technologies", and one from ``Mathematics and its Technologies''. While the first question was considered difficult by teachers due to the presence of distractors\footnote{Distractors are components of multiple-choice tests that can mislead test takers and make the incorrect alternative appears to be correct.}, the other two were graded as moderately difficult. By selecting questions from different areas and varying levels of difficulty, this research aims to ensure the generalizability of results to provide a thorough evaluation of this approach. Figure~\ref{fig:fewshot_examples_2022} presents the three few-shot examples (translated into English) that are used on the prompts in the same order as they appear.
Appendix~\ref{app:fewshot} shows the original questions in Portuguese.

\begin{figure}[!htb]
\sffamily
\tiny
\centering
\begin{tabular}{|
>{\columncolor[HTML]{ECF4FF}}p{0.9\columnwidth} |}
\hline
{\color[HTML]{343434} \textbf{Question 1:}}  \\
{\color[HTML]{343434} \textbf{Header:} Emotional urgency. If everything is for yesterday, if life engages a first gear and takes off, if there is no more time for strategic stops, we fatally fall into the addiction of wanting love to be equally resolved in a split second. We are in a hurry to hear ``I love you". We can't wait for the rules of coexistence to be established: are we boyfriends, hookups, married, lovers? Emotional urgency. A trap. We associate several words with LOVE: passion, romance, sex, adrenaline, palpitation. We forget, however, the word that makes this feeling possible: ``patience". Love without patience does not avenge. Love cannot be chewed and swallowed with emergency, desperate hunger. It's a meal that can last a lifetime. MEDEIROS, M. Available at: http://porumavidasimples.blogspot.com.br. Accessed on: 20 Aug. 2017 (adapted).} \\
{\color[HTML]{343434} \textbf{Statement:} In this opinion text, the linguistic marks reveal a relaxed situation with little formality, which is evidenced by the} \\
{\color[HTML]{343434} \textbf{Alternatives:}}     \\ 
{\color[HTML]{343434} A. impersonalization throughout the text, as in: ``if there is no more time''.} \\ 
{\color[HTML]{343434} B. construction of an atmosphere of urgency, in words like: ``hurry''.} \\ 
{\color[HTML]{343434} C. repetition of a certain syntactic structure, as in: ``If everything is for yesterday''.} \\ 
{\color[HTML]{343434} D. emphasis on the use of hyperbole, as in: ``a meal that can last a lifetime''.} \\ 
{\color[HTML]{343434} E. use of metaphors, as in: ``life engages a first gear and takes off''.} \\ 
{\color[HTML]{343434} \textbf{Answer:}} {\color[HTML]{00009B} \textit{E. use of metaphors, as in: ``life engages a first gear and takes off''.}}     \\ \\
\textbf{\#\#} \\
{\color[HTML]{343434} \textbf{Question 2:}}  \\
{\color[HTML]{343434} \textbf{Header:} Whenever the relevance of discourse comes into play, the issue becomes political by definition, as it is the discourse that makes man a political being. And everything men do, know or experience only makes sense to the extent that it can be discussed. There will perhaps be truths that lie beyond language and that may be of great relevance to man in the singular, that is, to man who, whatever he may be, is not a political being. But men in the plural, that is, the men who live and move and act in this world, can only experience the meaning of things by being able to speak and be intelligible to each other and to themselves. ARENDT, H. The human condition. Rio de Janeiro: University Forensics, 2004.} \\
{\color[HTML]{343434} \textbf{Statement:} In the excerpt, the philosopher Hannah Arendt shows the importance of language in the process of} \\
{\color[HTML]{343434} \textbf{Alternatives:}}     \\ 
{\color[HTML]{343434} A. understanding of culture.} \\ 
{\color[HTML]{343434} B. increased creativity.} \\ 
{\color[HTML]{343434} C. perception of individuality.} \\ 
{\color[HTML]{343434} D. improvement of technique.} \\ 
{\color[HTML]{343434} E. construction of sociability.} \\ 
{\color[HTML]{343434} \textbf{Answer:}} {\color[HTML]{00009B} \textit{E. construction of sociability.}}     \\ \\
\textbf{\#\#} \\
{\color[HTML]{343434} \textbf{Question 3:}}  \\
{\color[HTML]{343434} \textbf{Header:} A couple plans to build a swimming pool in the shape of a rectangular parallelepiped with a capacity of 90,000 L of water on their farm. The couple hired a construction company that presented five projects with different combinations of the internal dimensions of depth, width and length. The pool to be built will have the same ceramic coating on its walls and bottom, and the couple will choose the project that requires the smallest coating area. The internal dimensions of depth, width and length, respectively, for each of the projects are: project I: 1.8 m, 2.0 m and 25.0 m; project II: 2.0 m, 5.0 m and 9.0 m; project III: 1.0 m, 6.0 m and 15.0 m; project IV: 1.5 m, 15.0 m and 4.0 m; project V: 2.5 m, 3.0 m and 12.0 m.} \\
{\color[HTML]{343434} \textbf{Statement:} The project that the couple should choose will be the} \\
{\color[HTML]{343434} \textbf{Alternatives:}}     \\ 
{\color[HTML]{343434} A. I.} \\ 
{\color[HTML]{343434} B. II.} \\ 
{\color[HTML]{343434} C. III.} \\ 
{\color[HTML]{343434} D. IV.} \\ 
{\color[HTML]{343434} E. V.} \\ 
{\color[HTML]{343434} \textbf{Answer:}} {\color[HTML]{00009B} \textit{B. II.}}     \\ \hline
\end{tabular}
\caption{Questions from the ENEM 2022 exam used as few-shot examples.}
\label{fig:fewshot_examples_2022}
\end{figure}

For experiments on the ENEM 2022 dataset, when one of the three selected few-shot examples is evaluated, we exclude that example from the few-shot context and only use the remaining two examples as few-shot examples. 

\subsubsection{\textbf{Few-shot with CoT}}

Withal, this work investigates the enhancement of few-shot prompts with CoT techniques~\cite{DBLP:journals/corr/abs-2201-11903}. These prompts received additional sequences of explanatory steps, in order to evaluate the hypothesis that the models are able to decompose complex problems into smaller, more manageable parts; i.e., enabling the engagement of reasoning prior to compiling final results. This strategy would allow the model to potentially grasp abstract concepts and manipulate them, thereby facilitating the resolution of problems that require a deep understanding of underlying principles.

The CoT design we apply in this study starts with a brief summary of the statement. Afterward, the explanation points out the correct alternative and justifies it. This sequence can optionally be preceded by listing all alternatives that are likely correct, and followed by a sequence that excludes distractors. This strategy instructs the model to be aware of distractors. Also, the explanation ends with a justification for the remaining incorrect alternatives. This format is similar to the original proposal of CoT for common sense Q\&A~\cite{DBLP:journals/corr/abs-2201-11903}, except for the novel use of a section that intends to make the model aware of distractors.

To specify requirements and constraints that enable the model to reply with the expected structure, we have written the instruction at the beginning of the CoT prompts as follows:

\textit{Formulate a chain of explanations that allows you to answer the multiple-choice question below. Only one alternative is correct. Desired format: point out the alternatives that make sense, choose the CORRECT alternative and justify it, and finish justifying why the other alternatives are incorrect. Finish the explanation with ``Answer:'' followed by the alternative.}

The instruction was translated into English to facilitate the understanding of the method, but we use it in Portuguese when evaluating the models.

In addition to the instruction, each few-shot example of a prompt with CoT is modified as follows: after the set of alternatives, we replace ``response'' with ``explanation'' followed by the respective explanation. The explanation ends up with the alternative predicted as correct.

In our study, we utilize high-school teachers' discussions\footnote{Such as those from these links: \href{https://www.youtube.com/watch?v=gAvyffWAqxg}{1}, \href{https://g1.globo.com/educacao/enem/video/enem-2022-correcao-da-questao-de-filosofia-sobre-politica-e-linguagem-11122067.ghtml}{2}, \href{https://g1.globo.com/educacao/enem/video/enem-2022-correcao-da-questao-de-portugues-sobre-urgencia-emocional-11122141.ghtml}{3}, and \href{https://descomplica.com.br/gabarito-enem/questoes/2022/segundo-dia/o-projeto-que-o-casal-devera-escolher-sera-o/}{4}.} to formulate explanations for each few-shot example, as shown in Figure~\ref{fig:explanations_enem_2022}.
Appendix~\ref{app:cot_explanations} presents the original explanations in Portuguese.

\begin{figure}[!htb]
\sffamily
\tiny
\centering
\begin{tabular}{|
>{\columncolor[HTML]{ECF4FF}}p{0.9\columnwidth} |}
\hline
{\color[HTML]{343434} \textbf{Question 1:}}  \\
{\color[HTML]{343434} The text is written in a light, agile language, with little formality. In addition, it has figures of speech, such as metaphors and hyperboles, which are not mutually exclusive. In a sequential analysis of the alternatives, it would be possible to affirm that D) and E) are correct. However, looking in detail, it is noted that the expression ``use of metaphors'' proves to be more appropriate than ``emphasis on the use of hyperbole'', since, in order to state that the use of hyperbole was emphasized, the figure of speech should have appeared more often. This makes option E) more likely to be CORRECT. In addition, impersonality should not be pointed out as a mark of low formality. There is also an atmosphere of urgency, but that is criticized in the text that highlights the importance of patience and not haste. Finally, the syntactic structure is not systematically repeated throughout the text. Answer: E.} \\ \\ \hline
{\color[HTML]{343434} \textbf{Question 2:}} \\
{\color[HTML]{343434} Hannah Arendt argues in her work that we are political beings, in the proper sense of living in a polis, in a collective and social environment. And this sociability is only possible through discourse, language. Thus, we can conclude that language is an important tool for building sociability, and therefore alternative E) is CORRECT. Furthermore, it is not about understanding the culture, but the social relationship between people of that culture. Hannah also doesn't talk about increased creativity, nor does she talk about technique. Finally, language is used in something more collective and social, just the opposite of individuality. Answer: E.} \\ \\ \hline
{\color[HTML]{343434} \textbf{Question 3:}} \\
{\color[HTML]{343434} We must calculate the area of the four side faces and the area of the lower base (bottom of the pool) and add these areas to obtain the coating area. Therefore, calculating the coating area of each project, we have: Project I: A = 2 x 25 + 2 x 1.8 x (2 + 25) = 147.2; Project II: A = 9 x 5 + 2 x 2 x (9 + 5) = 101; Project III: A = 15 x 6 + 2 x 1 x (15 + 6) = 132; Project IV: A = 4 x 15 + 2 x 1.5 x (15 + 4) = 117; Project V: A = 3 x 12 + 2 x 2.5 x (3 + 12) = 111. Therefore, the project with the smallest coating area is the project II, therefore the correct answer is B. Answer: B.} \\ \\ \hline
\end{tabular}
\caption{Explanations for the three questions selected from the ENEM 2022 exam as few-shot examples.}
\label{fig:explanations_enem_2022}
\end{figure}

\section{Results}
\label{sec:experiments}

This section presents the results for the latest ENEM exam (2022) and for the ENEM Challenge dataset (2009-2017). We reinforce that the few-shot experiments used three examples from ENEM 2022 (see Section~\ref{sec:fewshot_examples}), but when evaluating one of those, we exclude it from the prompt and use only the remaining two as few-shot examples.

\subsection{ENEM 2022}
\label{sec:results_enem_2022}

\begin{figure*}[t]
\begin{subfigure}{0.5\textwidth}
\includegraphics[width=0.98\linewidth]{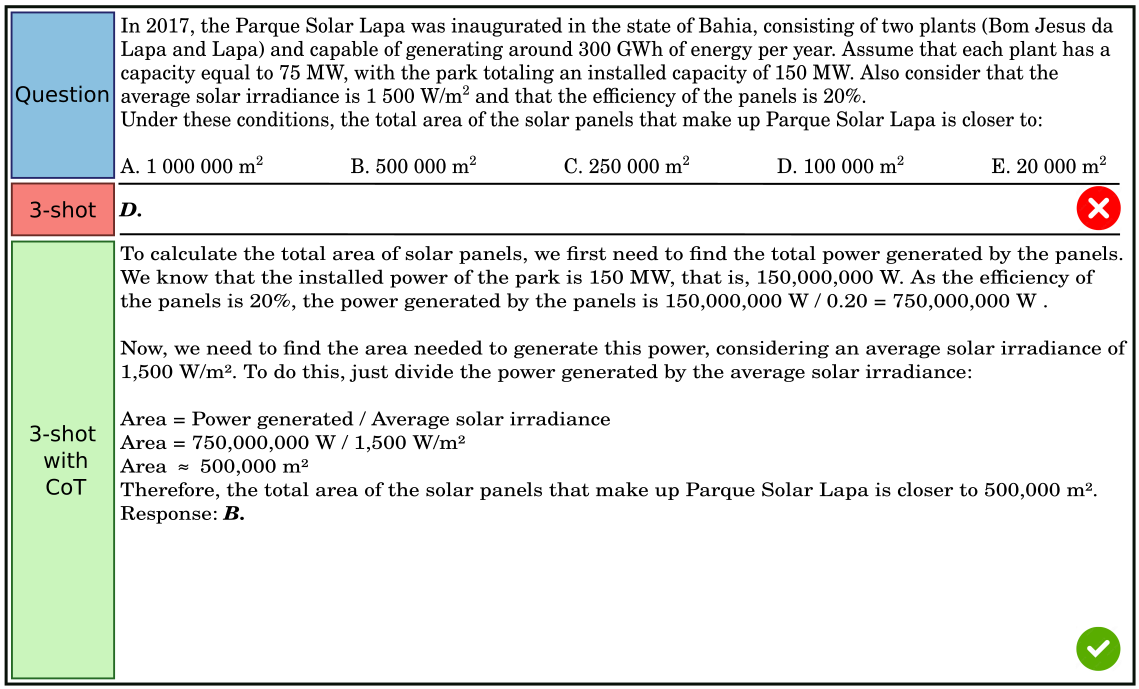} 
\caption{Question 134 of Natural Sciences}
\label{fig:subim1}
\end{subfigure}
\begin{subfigure}{0.5\textwidth}
\includegraphics[width=0.98\linewidth]{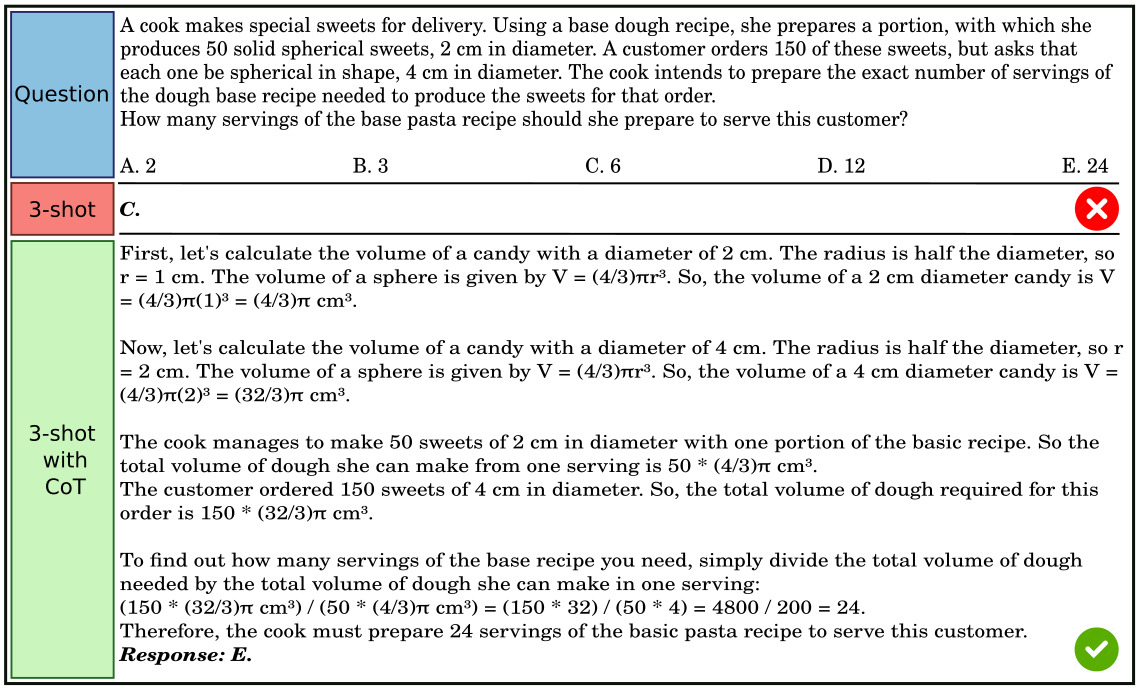}
\caption{Question 141 of Mathematics}
\label{fig:subim2}
\end{subfigure}
\caption{Comparison of answers generated by \texttt{gpt-4} with and without the Chain-of-Thought prompt. Questions are from ENEM 2022. The texts were translated into English for better comprehension.}
\label{fig:examples_gpt4_with_cot}
\end{figure*}

Table~\ref{tab:results_enem_2022} presents the results of the zero-shot, few-shot, and few-shot with CoT strategies applied on the latest ENEM exam.
In zero-shot experiments, the \texttt{gpt-4} model achieved an average accuracy of 79.66\%. It performed exceedingly well in human sciences questions (94.59\%), whereas its performance was subpar in mathematics questions (40.91\%). By utilizing few-shot prompts, the accuracy of the model increased by a mere one percentage point. However, we observed a substantial improvement of over 6 percentage points when using the few-shot CoT prompt, with mathematical questions showing the highest improvement (50.00\% $\rightarrow$ 72.73\%), followed by natural science questions (76.92\% $\rightarrow$ 88.46\%). This improvement is exemplified in Figure~\ref{fig:examples_gpt4_with_cot}, which shows questions requiring mathematical reasoning. The model becomes capable of solving complex problems when it begins to decompose them into smaller parts.

For \texttt{code-davinci-002} and \texttt{gpt-3.5-turbo}, the CoT prompt resulted in substantial improvements for mathematical questions. However, this was accompanied by a marked decline in accuracy in the remaining domains, ultimately leading to a decrease in overall accuracy when utilizing the CoT prompt.

\begin{table*}[!htb]
\centering
\caption{Results on ENEM 2022. Questions that require image comprehension were removed.}
\begin{tabular}{@{}lccccccccc@{}}
\toprule
\multicolumn{1}{c}{\multirow{2}{*}{Area}} & \multicolumn{3}{c}{code-davinci-002} & \multicolumn{3}{c}{gpt-3.5-turbo} & \multicolumn{3}{c}{gpt-4} \\ \cmidrule(lr){2-4} \cmidrule(lr){5-7} \cmidrule(lr){8-10}
\multicolumn{1}{c}{} &
  \begin{tabular}[c]{@{}c@{}}zero-shot\end{tabular} &
  \begin{tabular}[c]{@{}c@{}}three-shot\end{tabular} &
  \begin{tabular}[c]{@{}c@{}}three-shot \\with CoT\end{tabular} &
  \begin{tabular}[c]{@{}c@{}}zero-shot\end{tabular} &
  \begin{tabular}[c]{@{}c@{}}three-shot\end{tabular} &
  \begin{tabular}[c]{@{}c@{}}three-shot \\with CoT\end{tabular} &
  \begin{tabular}[c]{@{}c@{}}zero-shot\end{tabular} &
  \begin{tabular}[c]{@{}c@{}}three-shot\end{tabular} &
  \begin{tabular}[c]{@{}c@{}}three-shot \\with CoT\end{tabular} \\ \midrule
Languages and Codes   & 78.79       & 87.88       & 72.73       & 75.76       & 81.82       & 69.70       & 84.85       & 87.88       & 87.88       \\
Human Sciences     & 89.19       & 94.59       & 91.89       & 91.89       & 89.19       & 94.59       & 94.59       & 94.59       & 94.59       \\
Natural Sciences   & 69.23       & 61.54       & 53.85       & 73.08       & 84.62       & 65.38       & 84.62       & 76.92       & 88.46       \\
Mathematics          & 18.18       & 27.27       & 50.00       & 18.18       & 36.36       & 54.55       & 40.91       & 50.00       & 72.73       \\ \midrule
Total                                     & 68.64       & 72.88       & 70.34       & 69.49       & 76.27       & 73.73       & 79.66       & 80.51       & 87.29       \\ \bottomrule
\end{tabular}
\label{tab:results_enem_2022}
\end{table*}

\subsection{ENEM Challenge}
\label{sec:results_enem_2009-2017}

Table~\ref{tab:results_enem_2009-2017_2022} presents results for more models on the ENEM Challenge dataset\footnote{Two ENEM exams were held in 2016.}. Following Silveira et al~\cite{silveira2018enem}, we removed questions that require image comprehension, mathematical reasoning and chemical elements. The average zero-shot accuracy for the GPT-3.5 models spanned from 77.41\% to 82.88\%, whereas the \texttt{gpt-4} model exhibited a pronounced increase, attaining an accuracy of 94.56\%.

\begin{table*}[!htb]
\centering
\caption{Results on ENEM 2009-2017 and 2022 using zero-shot prompts. Questions that require image comprehension, mathematical reasoning, and chemical symbol understanding were removed.}
\begin{tabular}{@{}lccccccccccc|c@{}}
\toprule
Model            & 2009  & 2010  & 2011  & 2012  & 2013  & 2014  & 2015  & 2016\_1  & 2016\_2  & 2017 & avg. & 2022  \\ \midrule
Silveira et. al  & 26.96 & 31.37 & 33.85 & 31.34 & 30.33 & 27.01 & 29.96 & 28.42 & 27.95 & 26.68 & \textit{29.39} & - \\
text-davinci-002 & 74.16 & 77.45 & 80.21 & 78.26 & 82.35 & 82.76 & 80.90 & 77.66 & 74.19 & 79.78 & \textit{78.77} & 84.09\\
text-davinci-003 & 68.54 & 72.55 & 83.33 & 78.26 & 77.65 & 79.31 & 76.40 & 79.79 & 78.49 & 79.78 & \textit{77.41} & 79.55\\
code-davinci-002 & 83.15 & 82.35 & 87.50 & 81.52 & 80.00 & 87.36 & 84.27 & 75.53 & 81.72 & 85.39 & \textit{82.88} & 85.23 \\
gpt-3.5-turbo    & 78.65 & 77.45 & 85.42 & 84.78 & 74.12 & 82.76 & 84.27 & 85.11 & 76.34 & 84.27 & 
\textit{81.32} & 86.36 \\
gpt-4            & 95.51 & 92.16 & 97.92 & 98.91 & 91.76 & 96.55 & 96.63 & 92.55 & 89.25 & 94.38 & \textbf{\textit{94.56}} & 90.91     \\ \bottomrule      
\end{tabular}%
\label{tab:results_enem_2009-2017_2022}
\end{table*}

Additionally, we also investigated the zero-shot performance of the ENEM 2022 dataset using the same filtering criteria of the ENEM Challenge. Results are shown in the last column of Table~\ref{tab:results_enem_2009-2017_2022}. We observe that the GPT-3.5 models exhibited accuracies comparable to or surpassing the mean of the other ten examinations, thereby reducing the suspicion of possible data contamination and validating the results as the new state-of-the-art for the ENEM Challenge dataset. Conversely, the \texttt{gpt-4} reached a lower accuracy in the ENEM 2022 than the other exams, thus raising a concern that the answers to the 2009-2017 exams were, to some extent, memorized by the model during its training.

\section{Conclusion}

This study establishes GPT-4 as the new state-of-the-art model for tackling the ENEM challenge and provides insights into the effectiveness of various prompting strategies. The CoT prompt yielded significant improvements in terms of accuracy, while equipping the model with the ability to generate explanations to answers. This capability has the potential as an educational tool as it could enhance students' understanding of complex concepts and support their learning process by offering more transparent and informative responses to challenging questions. Additionally, we envision that LMs could have a substantial impact on the education sector, potentially leading to the adoption of AI-powered tools for psychometric analysis of exams, including predicting question difficulty, generating new exam items, and ultimately supporting the creation of adaptive test-suites.

\section{Future Work}

Building upon the findings of this study, several avenues for future research can be explored. First, we propose the development and evaluation of applications that leverage the capabilities of LMs for precision education, as outlined by \cite{luan2021review}. Such applications have the potential to enhance personalized learning experiences and improve educational outcomes.
More specifically, we suggest further investigation into the generation of questions within specific knowledge domains, as well as the calculation of difficulty levels of these questions. This line of research will provide valuable insights into the ability of LMs to create and adapt to varying degrees of complexity in the assessment tasks.

Lastly, we recommend the utilization of multimodal models\footnote{The multimodal feature of GPT-4 was not available to the public at the time of writing this paper.} and the extension of the ENEM evaluation to questions that require image comprehension. By incorporating these elements, researchers can explore the full range of the potential of AI models in addressing complex and diverse educational challenges.

\section{Acknowledgments}
\label{sec:acks}
We thank Thales R. Sales Almeida for providing the scrapping code to parse questions of the ENEM 2022 exam. This research was partially funded by \textit{Fundação de Amparo à Pesquisa do Estado de São Paulo} (FAPESP) (project id 2022/01640-2).

\bibliographystyle{unsrt}
\bibliography{main.bib}

\clearpage

\appendix

\section{Appendices}

\subsection{Few-shot examples}
\label{app:fewshot}

Figure~\ref{fig:fewshot_examples_2022_portuguese} depicts in Portuguese the three few-shot examples that are used to compose the prompt. We kept the examples in the same order as they appear in the few-shot context.

\begin{figure}[!htb]
\sffamily
\tiny
\centering
\begin{tabular}{|
>{\columncolor[HTML]{ECF4FF}}p{0.9\columnwidth} |}
\hline
{\color[HTML]{343434} \textbf{Questão 1:}}  \\
{\color[HTML]{343434} \textbf{Cabeçalho:} Urgência emocional. Se tudo é para ontem, se a vida engata uma primeira e sai em disparada, se não há mais tempo para paradas estratégicas, caímos fatalmente no vício de querer que os amores sejam igualmente resolvidos num átimo de segundo. Temos pressa para ouvir ``eu te amo''. Não vemos a hora de que fiquem estabelecidas as regras de convívio: somos namorados, ficantes, casados, amantes? Urgência emocional. Uma cilada. Associamos diversas palavras ao AMOR: paixão, romance, sexo, adrenalina, palpitação. Esquecemos, no entanto, da palavra que viabiliza esse sentimento: ``paciência''. Amor sem paciência não vinga. Amor não pode ser mastigado e engolido com emergência, com fome desesperada. É uma refeição que pode durar uma vida. MEDEIROS, M. Disponível em: http://porumavidasimples.blogspot.com.br. Acesso em: 20 ago. 2017 (adaptado).} \\
{\color[HTML]{343434} \textbf{Enunciado:} Nesse texto de opinião, as marcas linguísticas revelam uma situação distensa e de pouca formalidade, o que se evidencia pelo(a)} \\
{\color[HTML]{343434} \textbf{Alternativas:}}     \\ 
{\color[HTML]{343434} A. impessoalização ao longo do texto, como em: ``se não há mais tempo''.} \\ 
{\color[HTML]{343434} B. construção de uma atmosfera de urgência, em palavras como: ``pressa''.} \\ 
{\color[HTML]{343434} C. repetição de uma determinada estrutura sintática, como em: ``Se tudo é para ontem''.} \\ 
{\color[HTML]{343434} D. ênfase no emprego da hipérbole, como em: ``uma refeição que pode durar uma vida''.} \\ 
{\color[HTML]{343434} E. emprego de metáforas, como em: ``a vida engata uma primeira e sai em disparada''.} \\ 
{\color[HTML]{343434} \textbf{Resposta:}} {\color[HTML]{00009B} \textit{E. emprego de metáforas, como em: ``a vida engata uma primeira e sai em disparada''.}}     \\ \\
\textbf{\#\#} \\
{\color[HTML]{343434} \textbf{Questão 2:}}  \\
{\color[HTML]{343434} \textbf{Cabeçalho:} Sempre que a relevância do discurso entra em jogo, a questão torna-se política por definição, pois é o discurso que faz do homem um ser político. E tudo que os homens fazem, sabem ou experimentam só tem sentido na medida em que pode ser discutido. Haverá, talvez, verdades que ficam além da linguagem e que podem ser de grande relevância para o homem no singular, isto é, para o homem que, seja o que for, não é um ser político. Mas homens no plural, isto é, os homens que vivem e se movem e agem neste mundo, só podem experimentar o significado das coisas por poderem falar e ser inteligíveis entre si e consigo mesmos. ARENDT, H. A condição humana. Rio de Janeiro: Forense Universitária, 2004.} \\
{\color[HTML]{343434} \textbf{Enunciado:} No trecho, a filósofa Hannah Arendt mostra a importância da linguagem no processo de} \\
{\color[HTML]{343434} \textbf{Alternativas:}}     \\ 
{\color[HTML]{343434} A. entendimento da cultura.} \\ 
{\color[HTML]{343434} B. aumento da criatividade.} \\ 
{\color[HTML]{343434} C. percepção da individualidade.} \\ 
{\color[HTML]{343434} D. melhoria da técnica.} \\ 
{\color[HTML]{343434} E. construção da sociabilidade.} \\ 
{\color[HTML]{343434} \textbf{Resposta}} {\color[HTML]{00009B} \textit{E. construção da sociabilidade.}}     \\ \\
\textbf{\#\#} \\
{\color[HTML]{343434} \textbf{Questão 3:}}  \\
{\color[HTML]{343434} \textbf{Cabeçalho:} Um casal planeja construir em sua chácara uma piscina com o formato de um paralelepípedo reto retângulo com capacidade para 90 000 L de água. O casal contratou uma empresa de construções que apresentou cinco projetos com diferentes combinações nas dimensões internas de profundidade, largura e comprimento. A piscina a ser construída terá revestimento interno em suas paredes e fundo com uma mesma cerâmica, e o casal irá escolher o projeto que exija a menor área de revestimento. As dimensões internas de profundidade, largura e comprimento, respectivamente, para cada um dos projetos, são: projeto I: 1,8 m, 2,0 m e 25,0 m; projeto II: 2,0 m, 5,0 m e 9,0 m; projeto III: 1,0 m, 6,0 m e 15,0 m; projeto IV: 1,5 m, 15,0 m e 4,0 m; projeto V: 2,5 m, 3,0 m e 12,0 m.} \\
{\color[HTML]{343434} \textbf{Enunciado:} O projeto que o casal deverá escolher será o} \\
{\color[HTML]{343434} \textbf{Alternativas:}}     \\ 
{\color[HTML]{343434} A. I.} \\ 
{\color[HTML]{343434} B. II.} \\ 
{\color[HTML]{343434} C. III.} \\ 
{\color[HTML]{343434} D. IV.} \\ 
{\color[HTML]{343434} E. V.} \\ 
{\color[HTML]{343434} \textbf{Resposta:}} {\color[HTML]{00009B} \textit{B. II.}}     \\ \hline
\end{tabular}
\caption{Questions from the ENEM 2022 exam used as few-shot examples (in Portuguese).}
\label{fig:fewshot_examples_2022_portuguese}
\end{figure}

\subsection{CoT Explanations}
\label{app:cot_explanations}

Figure~\ref{fig:explanations_enem_2022_portuguese} presents in Portuguese the explanations formulated for the three few-shot examples. The explanations were created from contents of high-school teachers' discussions.

\begin{figure}[!htb]
\sffamily
\tiny
\centering
\begin{tabular}{|
>{\columncolor[HTML]{ECF4FF}}p{0.9\columnwidth} |}
\hline
{\color[HTML]{343434} \textbf{Questão 1:}}  \\
{\color[HTML]{343434} O texto é escrito em uma linguagem leve, ágil, e de pouca formalidade. Além disso, possui figuras de linguagem, como metáforas e hipérboles, que não são excludentes. Em uma análise sequencial das alternativas, daria para afirmar que D. e E. estão corretas. Entretanto, observando em detalhes, nota-se que a expressão "emprego de metáforas" mostra ser mais adequada do que "ênfase no emprego da hipérbole", visto que, para afirmarmos que o uso de hipérboles foi enfatizado, a figura de linguagem deveria ter aparecido mais vezes. Isso torna a alternativa E. mais provável de ser CORRETA. Além disso, impessoalização não deve ser apontada como marca de pouca formalidade. Existe também uma atmosfera de urgência, mas que é criticada no texto que destaca a importância da paciência e não da pressa. Por fim, a estrutura sintática não é repetida sistematicamente ao longo do texto. Resposta: E.} \\ \\ \hline
{\color[HTML]{343434} \textbf{Questão 2:}} \\
{\color[HTML]{343434} Hannah Arendt defende em sua obra que somos seres políticos, no sentido próprio de vivermos em pólis, em ambiente coletivo e social. E essa sociabilidade só é possível por meio do discurso, da linguagem. Desse modo, podemos concluir que a linguagem se apresenta como uma importante ferramenta para a construção da sociabilidade, e portanto a alternativa E. é a CORRETA. Além disso, não se trata do entendimento da cultura, mas da relação social entre as pessoas dessa cultura. Hannah também não fala sobre aumento de criatividade, tampouco sobre técnica. Por fim, a linguagem é utilizada em algo mais coletivo e social, justamente o oposto da individualidade. Resposta: E.} \\ \\ \hline
{\color[HTML]{343434} \textbf{Questão 3:}} \\
{\color[HTML]{343434} Devemos calcular a área das quatro faces laterais e a área da base inferior (fundo da piscina) e somar essas áreas para obter a área de revestimento. Logo, calculando a área de revestimento de cada projeto, temos: Projeto I: A = 2 x 25 + 2 x 1,8 x (2 + 25) = 147,2; Projeto II: A = 9 x 5 + 2 x 2 x (9 + 5) = 101; Projeto III: A = 15 x 6 + 2 x 1 x (15 + 6) = 132; Projeto IV: A = 4 x 15 + 2 x 1,5 x (15 + 4) = 117; Projeto V: A = 3 x 12 + 2 x 2,5 x (3 + 12) = 111. Logo, o projeto com menor área de revestimento, é o projeto II, portanto a resposta corrreta é B. Resposta: B.} \\ \\ \hline
\end{tabular}
\caption{Explanations for the three questions selected from the ENEM 2022 exam as few-shot examples (in Portuguese).}
\label{fig:explanations_enem_2022_portuguese}
\end{figure}

\end{document}